\documentclass[prl,twocolumn,twoside,showpacs,byrevtex,superscriptaddress]{revtex4}

\usepackage{currfile}

\usepackage{graphicx,epsfig,verbatim,enumerate}
\usepackage{amssymb,amsmath}
\usepackage{ifthen}

\usepackage{tikz-cd}
\usetikzlibrary{arrows}
\tikzset{
  commutative diagrams/.cd,
  arrow style=tikz,
  diagrams={>=space}}

\newboolean{twocolswitch}

\newcommand{\sindex}[1]{}
\newcommand{\nindex}[1]{}

\newcommand{\www}[1]{\url{#1}}

\usepackage{lettrine}

\usepackage{color}

\setboolean{twocolswitch}{true}

\begin{document}

\title{
  Is space is a word, too?

}

\author{
\firstname{Jake Ryland}
\surname{Williams}
}

\email{jw3477@drexel.edu}

\affiliation{
  Department of Information Science,
  College of Computing and Informatics,
  Drexel University,
  30 N. 33\textsuperscript{rd} St.
  Philadelphia, PA 19104}

\author{
\firstname{Giovanni C.}
\surname{Santia}
}

\email{gs495@drexel.edu}

\affiliation{
  Department of Information Science,
  College of Computing and Informatics,
  Drexel University,
  30 N. 33\textsuperscript{rd} St.
  Philadelphia, PA 19104}

\date{\today}

\begin{abstract}
  For words, rank-frequency distributions
have long been heralded for adherence to a
potentially-universal phenomenon known as Zipf's law.
The hypothetical form
of this empirical phenomenon
was refined by Ben\^{i}ot Mandelbrot
to that which is presently referred to
as the Zipf-Mandelbrot law.
Parallel to this,
Herbert Simon proposed a selection model
potentially explaining Zipf's law.
However, a significant dispute between Simon and Mandelbrot,
notable empirical exceptions,
and the lack of a strong empirical connection
between Simon's model and the Zipf-Mandelbrot law
have left the questions of
universality and mechanistic generation open.
We offer a resolution to these issues by exhibiting how
the dark matter of word segmentation,
i.e., space, punctuation, etc.,
connect the Zipf-Mandelbrot law
to Simon's mechanistic process.
This explains Mandelbrot's refinement
as no more than a fudge factor,
accommodating the effects of
the exclusion of the rank-frequency dark matter.
Thus, integrating these non-word objects
resolves a more-generalized rank-frequency law.
Since this relies upon the integration of space, etc.,
we find support for the hypothesis that
\emph{all} are generated by common processes,
indicating from a physical perspective
that space is a word, too.
  
\end{abstract}

\pacs{89.75.-Da, 02.50.Ey, 89.70.-a, 89.75.-Fb}

\maketitle

Where does Zipf's law come from?
This potentially universal linguistic phenomenon
is simple to express, but its origins have
a large diversity of explanations~\cite{zipf1949a,mandelbrot1953a,simon1955a,miller1957a,Li1992a,CorominasMurtra2015a}.
Its applicability
and the methods for its determination
remain subjects of discussion~\cite{Urzua2000,clauset2009b,piantadosi2013a,gerlach2013a},
which have coincided with observations of its variation~\cite{ferrericancho2005a}
and gross deviations~\cite{ferrericancho2001c}.
We have devoted significant effort to
understanding these breaches of universality;
the granularity of linguistic units (e.g., words vs. phrases)~\cite{williams2015a}
and the scales of corpora (e.g, documents vs. compilations)~\cite{williams2015b}
impact the measurable properties of Zipf's law.
Zipf's law has even been observed
to describe the frequencies of punctuation
along with words~\cite{kulig2017a};
surprisingly, it may describe whitespace, too.

One possible generation mechanism
also underpins the preferential attachment concept
from complex networks~\cite{barabasi1999a,krapivsky2001a},
and traces its roots to the study of evolution~\cite{yule1924a}.
This linguistic selection model
was proposed by Herbert Simon~\cite{simon1955a}
and has been the subject of our direct focus~\cite{dodds2017a},
exhibiting its prediction of
a scaling outlier at the first rank of Zipf's law.
This property presents an unforeseen opportunity
to explain the phenomenon.
If the scaling outlier is observed,
Simon's model has a smoking gun behind Zipf's law
as the only mechanism known to generate the first mover.
In the proceedings of this work,
we thus show how this preferential selection mechanism
is at the heart of Zipf's phenomenon.

For a text of $R$ distinct words: $w_1, \cdots, w_R$;
ranked, $r=1,\cdots,R$, descending, according to
their frequencies of occurrence:
$f(w_1) \geq \cdots f(w_R)$,
Zipf's law is a power-law distribution:
\begin{equation}
\label{eq:zipf}
\hat{f}\textsubscript{Z}(w_r) \propto r^{-\gamma};\:\:\:\:\: \gamma\geq0,
\end{equation}
which, in canonical form, has scaling exponent: $\gamma = 1$.
Empirical studies of rank-frequency distributions have exhibited
variation in $\gamma$~\cite{ferrericancho2005a,mehri2017a};
it is not uncommon for $\gamma$ to be greater or less than $1$,
but $\gamma$ is always greater than $0$
as a result of the rank ordering.
Following Zipf, Beno\^{i}t Mandelbrot observed a common tendency
of rank-frequency distributions to roll back,
falling beneath Zipf's law
at the highest ranks.
To suit, Mandelbrot
proposed a horizontal translation, $k$,
as a refinement of Zipf's law~\cite{mandelbrot1953a},
rooting it in the optimization of communication
against the cost of transmission.
This refinement by $k$ is commonly accepted
in the modern incarnation of Zipf's law
and is now empirically investigated along with $\gamma$~\cite{piantadosi2013a}.
The form of this Zipf-Mandelbrot (ZM) law is:
\begin{equation}
\label{eq:mandelbrot}
\hat{f}\textsubscript{ZM}(w_r) \propto (r+k)^{-\gamma}; \:\:\:\:\: \gamma\geq 0,k>-1.
\end{equation}

Seeking to understand the ZM law,
we are confronted with the question:
why would a translation of ranks
result in a more-accurate model?
For Zipf's or the ZM law, ranks 
more or less describe the number of
words of similar and more-severe frequency.
So, accepting a Mandelbrot translation of 2.75~\cite{piantadosi2013a},
should we conclude that there generally exist
between two and three mysterious words
of greater frequency than the largest observed?
If so, what are these words?
To approach these questions
we must first define what words are.

Approaches to defining words
for a rank-frequency study
include criteria like
things separated by space
and
accepted dictionary entries.
Studies handle case sensitivity differently,
and have even focused on the effects of lemmatization~\cite{corral2015a}.
In this work,
we choose not to modify case,
do not lemmatize,
and move forward with the criterion that
a word has at least one word character
to encompass a large number of word-like objects.
We define the set of word characters as alphanumerics,
their diacritic-modified forms, ligatures,
and any other characters not used as punctuation
from across all electronically-encoded natural languages
(although our focus will be on English).
Thus, we allow whitespace and all known punctuation characters
to act as word delimiters,
with the exception of contractions,
whose components' words are
separated in accordance to known rules,
such as:
$\{\text{don'tcha}\}\mapsto\{\text{do},\text{n't},\text{cha}\}$.

Returning to candidates for Mandelbrot's
two to three mysteriously-missing objects,
the process of defining words
has left a glaring possibility.
While semantic richness has made words
the historical focus of study,
this bias has hidden non-words---delimiters---away
as a linguistic dark matter.
Why is this conventional?
Space and punctuation, etc.,
may act as word delimiters,
but these objects have measurable frequencies, too.
Mechanistically, hand writing
relegates the placement of space
to a non-action,
but typewriting,
and now electronic encoding,
have made the insertion of space
into an active component of
the language-recording process.

\begin{figure}[t!]
  \includegraphics[width=0.5\textwidth, angle=0]{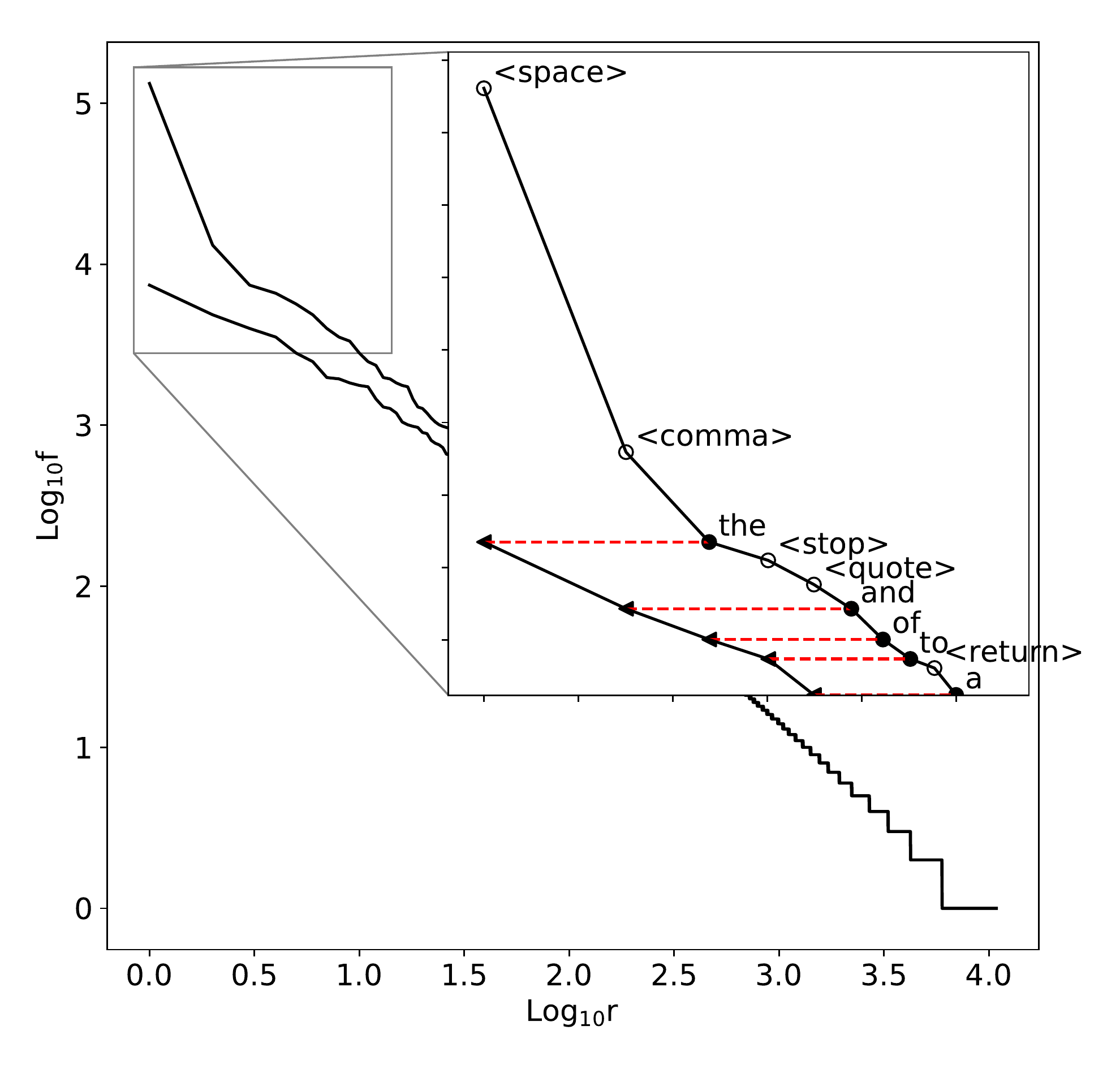}
  \caption{
    The translation of ranks for A Tale of Two Cities, by Charles Dickens.
    When both words and non-words are ranked by frequency (upper curve)
    the highest-frequency words emerge at rank 3---here,
    after space, and comma.
    When non-words are removed from the frequency ranking,
    all words are translated left (lower curve).
    This translation is highlighted in the inset zoom,
    where kept words move back along the red-dashed lines.
    Given the upper curve has a rank-frequency power law like Zipf's
    (Eq.~\ref{eq:zipf}),
    the deletion of non-words (open circles) most significantly
    affects the lowest-ranked/highest-frequency words (solid circles),
    rolling back to the solid triangles,
    potentially related to Mandelbrot's
    rank translation.
  }
  \label{fig:rankTranslation}
\end{figure}

So, how do our non-words integrate as
components of rank-frequency analyses?
Does the number of non-word objects ranking higher than the first word
explain Mandelbrot's shift?
Specific quantities actually vary from text to text,
but there is generally one non-word that takes rank 1 for English.
Perhaps unsurprisingly, this object is space,
which is often followed at rank 2 by comma.
An intuition for rank, frequency, and the word/non-word dichotomy
may be gained by observing how ranks map between
our integrated and the conventional rank-frequency representation.
Fig.~\ref{fig:rankTranslation} exhibits
how rank-frequency distributions roll back
when non-word objects are excluded.
Is this transition
related to Mandelbrot's $k$?
While the removal of non-words results
in a translation in the correct direction,
Mandelbrot's $k$ is defined constant.
This exists in contrast to
the shift by non-word deletion,
which is variable and increasing.
However, we may still experiment to
determine if the two are related.
Provided a relationship is uncovered,
Mandelbrot's $k$ may be no more than a
constant, fudge factor that
partially accounts for the missing non-words.

To understand any relationship between
the shift by non-word deletion and Mandelbrot's $k$,
we need a method for $k$'s measurement.
While the ZM law has a simple functional form,
power-law regression is notably difficult.
Accepted methodology for the determination of exponents
centers around maximum likelihood estimation~\cite{clauset2009b},
although a number of other methodologies exist~\cite{gabaix2012a,virkar2014a,corral2015a}.
However, we are not immediately interested in exponents or frequencies,
but instead, ranks.
Furthermore, optimization of $\gamma$ can obfuscate variation in $k$;
$\gamma$ is well-known to be
non-constant across ranks~\cite{ferrericancho2001c,montemurro2001a,gerlach2013a,williams2015b},
and is difficult to measure with precision~\cite{clauset2009b}.
For these reasons, we fix $\gamma = 1$
and switch to the rank domain for regression---the natural domain for $k$.
While fixing $\gamma$ imposes bias in the measurement of $k$,
it does so in a consistent way,
allowing us to precisely understand $k$'s variation,
which is all that is necessary for our study.

With $\gamma = 1$, a quotient of Zipf's law takes a convenient form:
$\hat{y}_\text{Z} = \hat{f}_\text{Z}(w_1)/\hat{f}_\text{Z}(w_r) = r$.
For the ZM law, the quotient becomes:
$\hat{y}_\text{ZM}(w_r) = (r+k)/(1+k)$.
Defining $y(w_r) = f(w_1)/f(w_r)$ for all $r$,
the sum of squared errors:
$\sum_{r = 1}^R(y(w_r) - \hat{y}_\text{ZM}(w_r))^2$
has analytic minimizer:
\begin{equation}
\label{eq:minimizer}
\hat{k} = \frac{\sum_{r = 1}^R(r - 1)^2}{\sum_{r = 1}^R(r - 1)(y(w_r) - 1)}.
\end{equation}
This regression is convenient to our study
for its analytic optimization.
Not only does this avoid the possibility of
multiple optimization minima,
but efficiencies in its computation 
allow observation of $\hat{k}$ as a function of $R$.

To distinguish traditional word-frequency distributions
from our space-integrated distributions
we express the latter
with $n$-ranks: $n=1,\cdots,N\geq R$;
subscript models and parameters, e.g.,
$\hat{f}_\text{S} \propto (n+k_\text{S})^{-\gamma_\text{S}}$
for space's integration;
and write $n_r$ to indicate the space-integrated rank
of the $r\textsuperscript{th}$ conventionally-ranked word.
For a text that perfectly conforms to
Zipf's law and our dark-matter hypothesis,
its space shift should be $\hat{k}_\textsubscript{S} = 0$,
making its ZM shift related to $n_1$,
potentially as $\hat{k}_\textsubscript{ZM} \approx n_1 - 1$.

\begin{figure}[t!]
  \includegraphics[width=0.5\textwidth, angle=0]{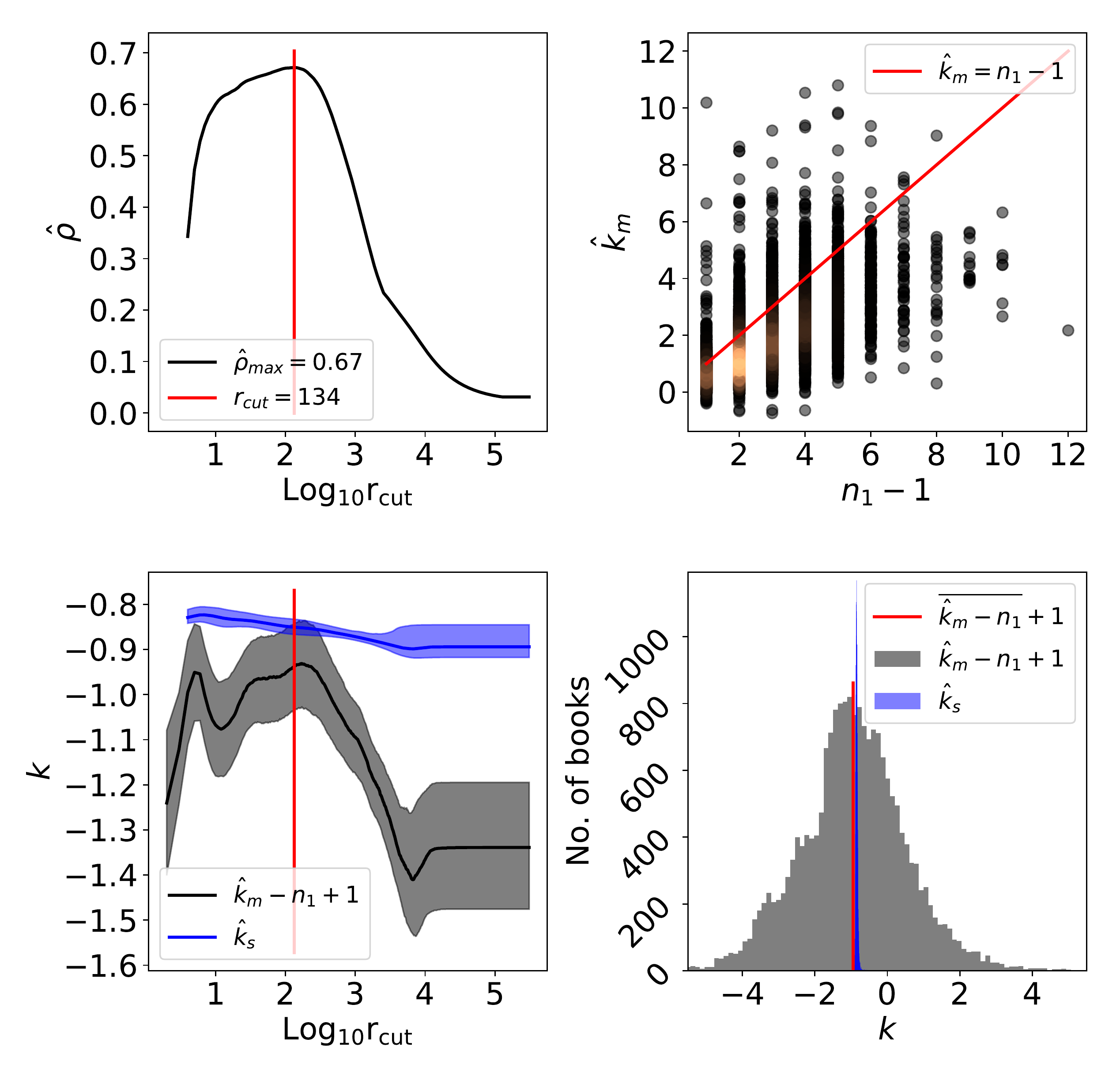}
  \caption{
    Top left: linear correlations between $\hat{k}_\text{ZM}$ and $n_1 - 1$
    as a function of increasing cutoffs, $r_\text{cut}$.
    The maximum value, $\hat{\rho}_\text{max} \approx 0.67$,
    is observed over the ranks:
    $r = 1, \cdots, r_\text{cut} = 134$,
    indicated by vertical red line.    
    Top right: scatter plot of $\hat{k}_\text{ZM}$ against $n_1 - 1$
    for the maximizing cutoff
    with density indicated by shade: low to high/black to copper.
    A perfect relationship is indicated by the diagonal red line,
    which appears offset from the most dense region by a value near $1$.
    Bottom left: median offset (black line), $\hat{k}_\text{ZM} - n_1 + 1$,
    as a function of increasing cutoffs, $r_\text{cut}$,
    centered in the offset distribution's middle decile (gray envelope).
    Offsets are compared to the median value for
    space-inclusive shifts (blue line),
    centered in the same distribution's interquartile range.
    The same correlation-optimal cutoff is indicated
    by a vertical red line.
    Bottom right: histograms of $\hat{k}_\text{ZM} - n_1 + 1$  (gray)
    and $\hat{k}_\text{S}$ (blue) for the 
    correlation-optimal cutoff,
    in addition to the average value of
    $\hat{k}_\text{ZM} - n_1 + 1$ (red line).
  }
  \label{fig:bookKs}
\end{figure}

To investigate if $\hat{k}_\textsubscript{ZM}$ and $n_1 - 1$ covary
we apply our rank regression to nearly $20,000$ English texts
from the Project Gutenberg eBooks collection.
We have used this data set in the past to
explain tail behaviors of rank-frequency distributions.
In this past study on text mixing~\cite{williams2015b}
we showed how the compilation of documents
into large, combined corpora leads to modulation
of Zipf's exponent, with more severe values of $\gamma > 1$ at high ranks.
The course of that study also led to our observation
that many of the eBooks are themselves mixed.
A notable example of our past study was the
Complete Historical Romances of Georg Ebers,
which is a compilation of over 100 short stories.
The existence of these texts 
has a capacity to throw off regression of
$\hat{k}_\textsubscript{ZM}$
in accommodation of the severe, mixing-derived exponents, $\gamma > 1$.
However, any true values of $k$
should be most apparent at smaller ranks~\cite{mandelbrot1953a},
i.e., $r\rightarrow 1$.
To separate the effects of
secondary scalings at high ranks, $r \gg 1$,
we explore an upper limit,
$r\textsubscript{cut} \leq R$,
on the number of terms in Eq.~\ref{eq:minimizer}.

The results of our experiment are non-trivial.
We $\rm{Log}$-space values of $r\textsubscript{cut}$
and measure the linear, Pearson correlations
between $\hat{k}_\textsubscript{ZM}$ and $n_1 - 1$.
While large cutoffs ($r\textsubscript{cut} \rightarrow R$)
exhibit little-to-no association,
we find a signal
that is maximized near $r\textsubscript{cut} = 100$,
for which $\hat{\rho}\approx 0.67$.
This is a strong correlation over nearly $20,000$ points,
and even though we have performed multiple comparisons
(676, for the $\rm{Log}$-spaced values of $r\textsubscript{cut}$),
the low p-values observed
(all p-values fell below $2\times10^{-5}$)
still test significant by
the most stringent, Bonferroni correction~\cite{dunn1961a}.
This scan can be observed
in Fig.~\ref{fig:bookKs} (top left),
showing how strong correlations coincide with small cutoffs
($r\textsubscript{cut} < 1,000$).
For the optimal cutoff ($r\textsubscript{cut} = 134$),
we exhibit a scatter plot (Fig.~\ref{fig:bookKs}, top right),
highlighting how $\hat{k}_\textsubscript{ZM}$
increases with $n_1 - 1$,
despite coarse, integer values of $n_1 - 1$.
While these two quantities increase together,
values of $n_1 - 1$ appear offset from $\hat{k}_\textsubscript{ZM}$
by a regular quantity near $1$.
This deviation does not appear to be random,
but is in fact in alignment with another quantity.
In the lower-left panel of Fig.~\ref{fig:bookKs},
variation (middle $10^\text{th}$ percentile, gray envelope) in the deviations,
$\hat{k}_\textsubscript{ZM} - n_1 + 1$,
is presented alongside
variation (middle $50^\text{th}$ percentile, blue envelope) in the shift
regressed from the space-inclusive frequency distribution,
$\hat{k}_\textsubscript{S}$.
Appearing much more stable than $\hat{k}_\textsubscript{ZM}$,
values of $\hat{k}_\textsubscript{S}$ nearly align
with the deviations at roughly the optimal cutoff.
\emph{Thus,
$\hat{k}_\textsubscript{ZM}$ is embedded with information
that resolves the parameterization of the space-inclusive distribution,
despite having no direct information on the presence of the excluded non-word objects!}
This result can be seen in the lower-right panel
of Fig.~\ref{fig:bookKs}, showing histograms of
$\hat{k}_\textsubscript{S}$ and $\hat{k}_\textsubscript{ZM} - n_1 + 1$,
where the variation in $\hat{k}_\textsubscript{S}$
is quite constrained, appearing like a vertical line.

While we have uncovered a strong signal relating
the inclusion of non-words to Mandelbrot's shift,
this signal has appeared in a surprising manner.
The space-inclusive distributions did not
exhibit shift values close to zero,
but instead extraordinarily regularly
near a different value, $\hat{k}_\textsubscript{S} \approx -0.85$.
The variation in this parameter is in fact
far more narrow than that found for Zipf's exponent~\cite{ferrericancho2005a,williams2015a,mehri2017a}.
Zipf's  scaling exponent, $\gamma$,
is prone to mixing effects,
which we have seen here to impact the
space-exclusive shift ($\hat{k}_\textsubscript{ZM}$)
when $r\textsubscript{cut}$ is roughly larger than $1000$.
What makes $\hat{k}_\textsubscript{S}$
so much more robust than $\hat{k}_\textsubscript{ZM}$?
At rank $n = 1$, space is extraordinarily dominant
in the frequency distributions, which is no surprise,
since it acts as the primary word delimiter for English.
More interestingly, though, is the range in which the
very-regular, negative value that $\hat{k}_\textsubscript{S}$ assumes.
Instead of describing a roll-back,
$\hat{k}_\textsubscript{S}$ describes
the opposite behavior---a disproportionately
large quantity of high-frequency non-words.
Unlike Mandelbrot's roll-back,
disproportionately high frequencies
\emph{are} actually predicted at these ranks
by a generative mechanism.

The rank-frequency empiricism of the $20\textsuperscript{th}$ century
was complimented by a stochastic model of language generation,
defined by Herbert Simon~\cite{simon1955a}.
The goal of this endeavor was to go beyond observation
of linguistic regularity to a mechanistic understanding
of its production.
Simon's model relied on a fixed word-innovation rate: $\alpha$.
With each time step,
either a new word is drawn with probability $\alpha$,
or an old word is drawn with probability $\theta = 1 - \alpha$.
The choice of old words is based on an independence assumption:
old words are drawn in proportion to their current frequencies.
While Simon's original analysis of this model
identified a power-law frequency distribution
of scaling exponent $\theta$,
our recent work with this model~\cite{dodds2017a}
identified how as $\theta\rightarrow 1$,
the frequency of the rank-$1$ word
becomes disproportionately large.
However, the disproportionate growth of
this rank-$1$ outlier is not encoded
by a piecewise functional form for frequency,
but as we will show,
a \emph{negative translation} that is most impactive at small ranks.
Thus, we now show how $\hat{k}_\textsubscript{S}$
is analytically predicted by Simon's model,
making space the run away, first mover
of English language.

To exhibit Simon's negative translation,
we must define $M_j: j=1,\cdots,N$, as the step at which 
the $j\textsuperscript{th}$ word is innovated.
Intuitively, $M_j$ is the number of words
produced by the time the $j\textsuperscript{th}$ unique word appears.
Previously~\cite{dodds2017a}, we determined
the analytic form for the resulting frequency distribution:
\begin{equation}
\label{eq:simonExact}
\hat{f}(w_j) = \frac{B(M_j,\theta)}{B(M_N,\theta)}\approx\left(\frac{M_j}{M_N}\right)^{-\theta}.
\end{equation}
Using this, our piecewise ansatz for $M_j$:
$M_1 = 1$; with $\langle M_j \rangle = (j-1)/\alpha$, for all $j > 1$
exhibited a first mover's disproportion,
since the rank-1 word's frequency
was not damped by $\alpha$.
In truth, this is \emph{not}
the real reason for the first mover's dominance
in Simon's model.
While near truth,
our ansatz misses a key analytic feature
that explains the first mover
via a negative translation.

For any $j\geq 1$,
$\langle M_{j+1} - M_j\rangle$
is the expectation of a geometric distribution
with success probability $\alpha$, i.e.,
$\langle M_{j+1} - M_j\rangle = \alpha^{-1}$.
Since one can show separation:
$\langle M_{j+1} - M_j\rangle = \langle M_{j+1}\rangle - \langle M_j\rangle$,
the resulting recursion equation:
$\langle M_{j+1}\rangle = \langle M_j \rangle + \alpha^{-1}$
provides:
\begin{equation}
\label{eq:Mestimate}
\langle M_j \rangle
= \frac{\alpha + j - 1}{\alpha}
= \frac{j - \theta}{\alpha}
.
\end{equation}
While the first form in Eq.~\ref{eq:Mestimate} exhibits the accuracy
of our old ansatz,
the latter will provide the most succinct
and illustrative expression of frequency.

Approximation of $M_j$
by $\langle M_j\rangle = (j - \theta)/\alpha$
into both the numerator and denominator of
Eq.~\ref{eq:simonExact} 
renders our refinement of
the Simon model's
analytic frequencies:
\begin{equation}
\label{eq:simon}
\hat{f}(w_j)
 \approx  \left(\frac{j - \theta}{N - \theta}\right)^{-\theta}.
\end{equation}
This unifies the Simon model's frequencies
into a common expression.
For $j = 1$, the negative shift by $\theta$ leaves
$j - \theta = \alpha < 1$ in the numerator,
whose reciprocal ($\hat{f}$ is hyperbolic) can be large.
This only occurs for the $j=1$, first mover,
and since estimates \cite{dodds2017a} place $\theta$ near $1$,
the entailed scaling outlier aligns with our
empirical observations presented in Fig.~\ref{fig:bookKs}.
Thus, our empirical connection between
Mandelbrot's $k$ and the exclusion of non-words
is simultaneously a smoking gun for Simon's model
as a primary mechanism for the generation of language
(English, at least), i.e., our measurement of $\hat{k}_\text{S}$
tracks $\theta$, though not exactly, since we have forced $\gamma = 1$.

So, what of languages other than English?
Many utilize non-word delimiters similarly,
and perhaps exhibit the space/first mover phenomenon.
However, word segmentation is less trivial
for a number of east-Asian languages,
with space infrequent, in a diminished role.
If Simon's model applies to these,
space's sparsity as a delimiter
and the absence of a dominant first mover
should be apparent in
values of $\hat{k}_\text{S}$ closer to zero
(though still negative).
This is an important avenue for study,
and requires adequate word-segmentation tools,
but is not the only question left open.
Zipf studied other domains;
e.g., do analogs for segmentation
and dark matter exist and what would they mean
for Zipfian city sizes~\cite{zipf1949a}?

Perhaps more immediately important is
how our results may
impact the language processing community.
Excluding space from frequency analysis
may be equivalent to
ignoring the single most massive celestial object
from a model of planetary motion.
Could space's inclusion have positive
effects on the engineering side?
Recently, we explored this possibility~\cite{williams2017a}.
Going beyond the incorporation of space, etc.,
we centered non-words as features in an algorithm.
This algorithm is now state of the art
for multiword expressions segmentation,
thus an example for the practical integration of non-words.
Furthermore, while many machine learning algorithms
exclude stop words in preprocessing,
Pennebaker~\cite{Pennebaker2011a} exhibited how variation of
pronouns and other high-frequency words
is predictive of a variety of
social and psychological characteristics.
Regardless,
this study provides a novel understanding of language,
both in terms of empirical laws,
and the nature of its formation.

The authors thank Sharon Williams
for her thoughtful discussions,
and gratefully acknowledge
research support from
Drexel University's
Department of Information Science
and College of Computing and Informatics.

\end{document}